\documentclass[pdflatex,sn-nature]{sn-jnl}

\usepackage{tabularx}
\usepackage{graphicx}%
\usepackage{multirow}%
\usepackage{amsmath,amssymb,amsfonts}%
\usepackage{amsthm}%
\usepackage{mathrsfs}%
\usepackage[title]{appendix}%
\usepackage{xcolor}%
\usepackage{textcomp}%
\usepackage{manyfoot}%
\usepackage{booktabs}%
\usepackage{algorithm}%
\usepackage{algorithmicx}%
\usepackage{algpseudocode}%
\usepackage{listings}%
\usepackage{hyperref}
\usepackage{makecell}

\theoremstyle{thmstyleone}%
\theoremstyle{thmstyletwo}%

\theoremstyle{thmstylethree}%

\begin{document}

\title[Article Title]{
\centering
In vivo feasibility study of humanoid robots
in surgery}


\author*[1]{\fnm{Zekai} \sur{Liang}}\email{z9liang@ucsd.edu}

\author[2]{\fnm{Nikita} \sur{Thareja}}\email{nthareja@health.ucsd.edu}

\author[1]{\fnm{Peihan} \sur{Zhang}}\email{pez004@ucsd.edu}

\author[1]{\fnm{Calvin} \sur{Joyce}}\email{cajoyce@ucsd.edu}

\author[1]{\fnm{Soofiyan} \sur{Atar}}\email{satar@ucsd.edu}

\author[1]{\fnm{Florian} \sur{Richter}}\email{frichter1995@gmail.com}

\author[2]{\fnm{Garth} \sur{Jacobsen}}\email{gjacobsen@health.ucsd.edu}

\author[2]{\fnm{Shanglei} \sur{Liu}}\email{s5liu@health.ucsd.edu}

\author[2]{\fnm{Ryan} \sur{Broderick}}\email{rbroderick@health.ucsd.edu}

\author[1]{\fnm{Michael} \sur{Yip}}\email{yip@ucsd.edu}

\affil[1]{\orgdiv{Jacobs School of Engineering}, \orgname{UC San Diego}, \orgaddress{\street{9500 Gilman Drive}, \city{La Jolla}, \postcode{92093}, \state{CA}, \country{USA}}}

\affil[2]{\orgdiv{Department of Surgery}, \orgname{UC San Diego}, \orgaddress{\street{9300 Campus Point Drive}, \city{La Jolla}, \postcode{92037}, \state{CA}, \country{USA}}}



\abstract{Recent advances across actuation, control, and learning have rapidly pushed humanoid robots from a distant vision toward near-term real-world deployment. Healthcare is a particularly pressing domain, where staffing shortages and increasing care demand are widening the gap between clinical workload and the availability of skilled labor. While current automation has largely focused on digital and logistical tasks, much of hospital work remains embodied, requiring mobility, manipulation, and safe interaction in human-designed environments. Humanoid form factors offer unique potential, particularly when used to assist with surgical tasks.
Traditionally, robotic systems for surgery are purpose-built platforms such as Intuitive Surgical's da Vinci Surgical System,  and it remains unclear how close current humanoid systems are to meeting the precision, control, and safety requirements of minimally invasive surgery.
In this work, we present a systematic evaluation of contemporary humanoid technology for laparoscopic surgical tasks. We develop a humanoid-based laparoscopic teleoperation framework using general-purpose instruments and assess its capabilities through benchtop characterization, dry-lab user studies spanning diverse surgical experience levels, and in-vivo porcine studies. Across these evaluations, we quantify technical feasibility, task performance, and clinical readiness relative to established surgical platforms. Together, our study provides an evidence-based assessment of the current capabilities and limitations of humanoids for surgical applications, highlighting both their promise and the key technical challenges that must be addressed before clinical deployment.  \textbf{Project website:} \url{https://humanoid-surgeon.github.io/}.
}

\maketitle

\section{Introduction}\label{sec1}

Over the past few years, human-like robots capable of walking, manipulating objects, and operating in everyday environments have moved from a distant vision toward an imminent reality. Humanoids are especially well-positioned for real-world deployment because their human-like morphology is compatible with spaces, tools, and workflows designed for people. This shift has been driven by sustained advances across the humanoid technology stack, particularly in actuation, control, and learning.
For example, modern humanoids increasingly adopt high-torque, low-impedance electric actuators for smooth and fast manipulation \cite{wensing2017proprioceptive, grimminger2020open, chignoli2021humanoid, semasinghe2025design}.
Meanwhile, on the software side, advances in whole-body model-predictive control \cite{khazoom2024tailoring,zhang2025whole, belvedere2024joint, ishihara2024hierarchical, vitor2025imitation}, reinforcement-learning-based locomotion policies \cite{radosavovic2024real,radosavovic2024learning,he2025viral}, and foundation-model driven perception and vision-language-action policies \cite{kim2024openvla,belkhale2024rt,wen2025tinyvla,zheng2024tracevla,o2024open} have dramatically improved balance robustness, contact-rich manipulation, and autonomy in unstructured environments. Beyond technical advances, the rapid development of humanoid robotics is increasingly reflected in growing industrial momentum and early commercialization efforts \cite{figure2025seriesc,bloomberg20251x}. These trends suggest that humanoids may soon be capable of performing labor-intensive human tasks at scale.

Healthcare represents a particularly critical application domain, where rising care demands and persistent workforce shortages place increasing strain on clinical systems \cite{who2023healthworkforce,who2024burnout,who_ageing_health_2025}. While existing hospital robots primarily support logistics and administrative tasks, extending robotic assistance to clinical workflows requires embodied interaction, dexterous manipulation, and safe operation in human-centered environments. Early deployments such as hospital logistics robots demonstrate the feasibility of robotic assistance in healthcare settings \cite{robotreport2025_moxi}. However, most existing systems focus on routine support tasks rather than safety-critical clinical procedures.

Beyond support and logistics, a particularly high-impact and safety-critical question is whether humanoid robots can assist with surgical tasks.  Indeed, the idea of humanoids performing surgery has captured public attention, fueled by high-profile claims of “superhuman surgeries” \cite{bruce2025musk_superhuman_surgeries}.
Prior work has extensively explored teleoperated robotic systems for minimally invasive surgery, most notably through purpose-built platforms such as the da Vinci Surgical System and related research platforms \cite{lanfranco2004robotic, taylor2016medical}. 
To reduce reliance on mechanically regulated remote center-of-motion (RCM) mechanisms, general-purpose manipulators have also been employed to maintain laparoscopic port constraints through kinematic control strategies \cite{marinho2014programmable, marinho2019dynamic, nasiri2024admittance, davila2024real}.
However, these systems still rely on robot-specific laparoscopic instruments \cite{hagn2010dlr, hagn2008dlr}, including those used in Intuitive's da Vinci Surgical System.
In contrast, Humanoid robots, with human-like kinematics and dexterous upper limbs, offer a potential pathway to directly manipulate surgical tools designed for human use.
Nevertheless, it remains unclear how close contemporary general-purpose humanoid robots are to meeting the precision, stability, and safety requirements of surgical procedures.

While purpose-built surgical platforms such as Intuitive Surgical's da Vinci Surgical System have demonstrated exceptional performance and widespread clinical adoption through years of industrial refinement, studies note that most existing operating rooms (ORs) were originally designed for manual surgery, and robotic-assisted surgery (RAS) often requires additional space and coordination for equipment positioning, robot docking, and intraoperative maneuvering \cite{cofran2022barriers,kanji2021room}. 
Humanoid robots, by contrast, share a similar form factor with humans, potentially offering practical deployment advantages in operating rooms. Their human-like morphology allows them to operate within existing surgical infrastructure and interact directly with standard manual medical equipment. 
Although the comparatively lower price tags of current humanoid platforms may partly reflect differences in regulatory, liability, and commercialization pathways compared with established surgical robots, humanoid systems offer the potential for versatile task deployment. Beyond minimally invasive surgery, humanoid robots capable of dexterous manipulation may support a broader range of clinical activities, including patient interaction \cite{ozturkcan2022humanoid, hernandez2024compassionate}, bedside assistance \cite{mlakar2025facilitating}, and hospital logistics \cite{haddadin2016physical,yang2018grand}. 
Such multi-purpose capabilities may improve resource utilization and reduce capital, storage, and operational demands in resource-constrained healthcare environments.

The recent Surgie study~\cite{atar2025humanoids} demonstrated that general-purpose platforms can be teleoperated to perform dexterous medical tasks under controlled conditions.
Although no surgical procedures were performed, the study suggests a plausible pathway from service-oriented hospital tasks toward minimally invasive surgical workflows. Yet a central question remains: can humanoid robots perform surgery safely? Bridging from controlled, teleoperated dexterous tasks to surgical capability requires levels of stability, precision, and safety assurance that no humanoid platform has yet demonstrated in clinical environments ~\cite{battaglia2021rethinking}.


In this work, we systematically evaluate the capabilities and limitations of contemporary humanoid platforms for surgical tasks by developing a humanoid-based laparoscopic teleoperation framework using general-purpose instruments. Laparoscopy is chosen as a representative benchmark due to its widespread clinical adoption and the availability of established robotic systems such as Intuitive’s da Vinci Surgical System.
We develop a teleoperation framework that enables a humanoid robot to perform laparoscopic manipulation using manual wristed instruments via a Master Tool Manipulator (MTM). Leveraging stereo vision and manual laparoscopic tools, the proposed framework maps intuitive human hand motions to scaled instrument control, enabling key capabilities of conventional minimally invasive surgical (MIS) robots.
We initially assess system performance through benchtop experiments and dry-lab studies, and further validate the framework in the first humanoid-based in-vivo porcine study in which the teleoperated humanoid completes two cholecystectomies.
Together, these results clarify how close current humanoids are to supporting laparoscopic surgical workflows and identify the key technical and safety gaps that must be addressed before broader clinical deployment.

\begin{figure}[!htbp] 
	\centering
	\includegraphics[width=0.95\textwidth]{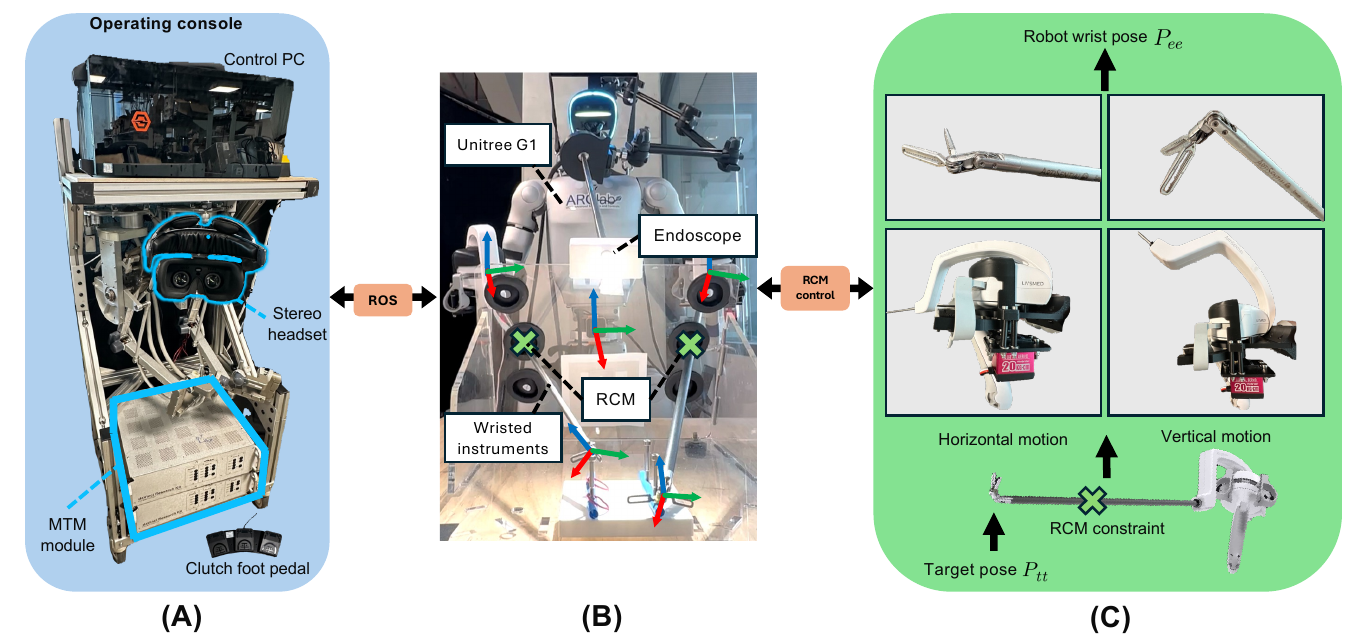} 

	\caption{\textbf{Teleoperation system of the surgical humanoid platform.}
		(\textbf{A}) The mobile operating console consists of a control PC, a stereo headset display, and a Master Tool Manipulator (MTM) module with a foot pedal. (\textbf{B}) Humanoid performs laparoscopic operations with manually-wristed instruments. (\textbf{C}) Instruments are mounted using a custom tool swapping mechanism to attach to the robot wrist, with gripper tools including a single active pinch degree of freedom on a servomotor to provide closing action on the grippers. End-effector orientation and positioning are accomplished under an extended kinematic chain under a remote center-of-motion (RCM) constraint. Communication interfaces between the mobile operating console and the humanoid robot are done via network control under the Robot Operating System (ROS)~\cite{doi:10.1126/scirobotics.abm6074}.}
	\label{fig:system} 
\end{figure}

\section{Results}\label{sec2}
The proposed framework is illustrated in Fig.~\ref{fig:system}, where a humanoid robot is teleoperated to manipulate manual laparoscopic instruments. A control console equipped with a stereo headset displaying the endoscopic view enables intuitive tool control. A non-trivial inverse-mapping control algorithm is developed to map tool-tip motion to robot wrist movements. The physical RCM is localized using ArUco markers detected by the robot’s front head camera. Detailed system design and technical specifications are provided in the Methods section.

\subsection{Benchtop experimentation}

To initially assess the capability of humanoid platforms for laparoscopic workflows, we perform controlled benchtop experiments using the Unitree G1 humanoid, an economical and commercially available system representative of contemporary general-purpose humanoids. These experiments establish a baseline performance level and identify technical limitations before in vivo evaluation. We evaluate workspace accessibility and maneuverability, and quantify teleoperation precision with manually held tools, which are critical capabilities for laparoscopic operation.

\subsubsection{Kinematic \& workspace analysis}

Laparoscopic surgery requires instruments to pass through a small trocar while maintaining a fixed pivot, known as the Remote Center of Motion (RCM). Under this constraint, the instrument tip typically sweeps a conical workspace \cite{wilson2010evaluating,degraag2020optimal}. Prior MIS design studies indicate that cone vertex angles of roughly $60^\circ$–$90^\circ$ and port-to-target distances of 10–20 cm are sufficient to support most laparoscopic tasks while preserving dexterity and visualization \cite{afshar2020optimal,degraag2020optimal}.


Conventional surgical robots satisfy this requirement through a mechanically constrained RCM aligned to the port during operating-room setup\cite{freschi2013technical,intuitive2021davinxi_orstaff}. In contrast, general-purpose humanoids lack mechanically constrained RCM and require a safe pivot through perception and control while using manual laparoscopic instruments, as shown in Fig.~\ref{fig:benchtop}(A). Consequently, their reachable surgical workspace greatly depends on the RCM placement with respect to the robot base.
We analyze the reachable instrument workspace as a function of the RCM placement and compare it with workspace specifications commonly used in robotic laparoscopy to assess whether the humanoid platform meets these kinematic requirements.

Two representative reachability examples for the humanoid are shown in Fig.~\ref{fig:benchtop}(B), which demonstrates how trocar placement can impact the reachable space inside a patient's body. For each case, we sample a dense grid of candidate robot hand positions within a predefined region of interest, and evaluate the feasibility using an inverse-kinematics (IK) solver subject to joint-limit and collision constraints. Feasible solutions define the humanoid hand workspace (green points). The corresponding surgical tool-tip positions are obtained via the forward kinematics of the extended kinematic chain, yielding the reachable tool workspace (blue points).

 Multiple RCM locations are sampled to analytically examine how insertion placement affects the humanoid's reachability in real surgical setups. The visualized reachable region is approximately cone-shaped, with both volume and dominant approach direction modulated by the RCM position. This dependence is further quantified on the $y$–$z$ plane, as shown in Fig.~\ref{fig:benchtop}(C): at a fixed height, tighter kinematic constraints reduce workspace volume when the RCM is moved closer to the humanoid torso.
As a benchmark, the dVRK workspace is modeled as a spherical sector with a 100$^\circ$ apex angle and 20 cm radius, reaching a total volume of 5985 cm$^3$ \cite{kazanzides2014open}. In comparison, the humanoid achieves comparable workspace for a subset of RCM placements, indicating that standard trocar placements need to be adapted for the humanoid before surgical deployments.

\subsubsection{Motion accuracy evaluation}
Motion accuracy is a key determinant for safely and effectively manipulating delicate anatomical structures during laparoscopic surgery. Prior studies \cite{richter2019motion, lu2022adaptive} have identified common sources of error in conventional teleoperation systems, including communication latency, kinematic inaccuracies, and human–robot interaction dynamics. In the proposed humanoid laparoscopic framework, manual laparoscopic instruments are incorporated as an extended kinematic chain with more complex control loops. Furthermore, because the commercial instruments used in this study are not publicly documented, the geometric parameters of the extended kinematic chain must be obtained through manual measurement and modeling, introducing potential calibration errors. To quantify and evaluate the motion accuracy of the system, we approach this from both system latency and command-execution tracking perspectives.


\textbf{System latency}: We characterize the system latency in terms of follower-leader latency and control loop latency. The follower-leader latency, defined as the delay between operator hand motion and robot end-effector motion, was measured as  $\sim156$ms using a high frame rate (120 FPS) external camera. The control loop latency measured on the control PC is approximately $24$ ms, within which the inverse kinematics computation requires approximately $11$ ms per cycle. Recent work \cite{xiong2026extremcontrol} reports that existing humanoid teleoperation systems commonly exhibit end-to-end latencies on the order of hundreds of milliseconds, suggesting that such platforms remain under active development and that further improvements are needed.  Prior studies indicate that latency below 150\,ms is generally considered desirable for conventional surgical robotic systems \cite{ichihara2025quantifying}.

\begin{figure}[!htbp] 
	\centering
	\includegraphics[width=\textwidth]{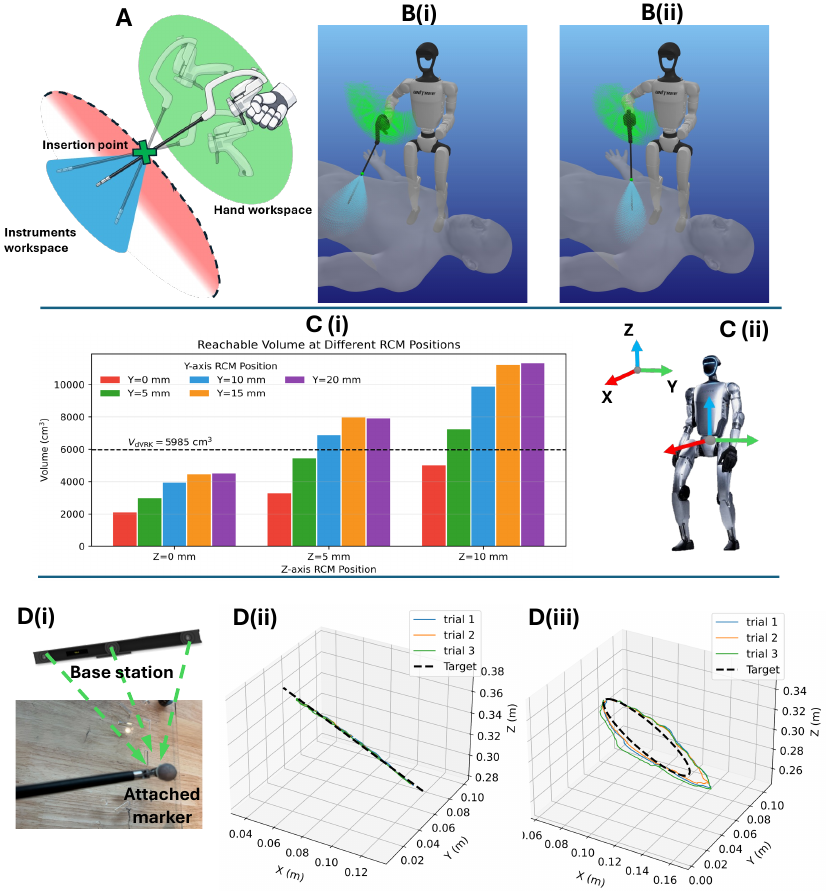} 
	\caption{\textbf{Results of benchtop examination.}
		\textbf{(A)} Hand workspace (green) and corresponding instrument workspace (blue) given a certain insertion position.
        \textbf{(B)(i, ii)} Instrument workspaces for different trocar placements with respect to the robot base, forming approximately cone-shaped reachable regions with varying volume and orientation.
        \textbf{(C)(i)} Cross-sections of the reachable workspace on the $y$--$z$ plane, showing reduced workspace as the RCM moves closer to the humanoid torso. \textbf{(ii)} Robot base frame position is at mid torso. \textbf{(D)} \textbf{(i)} Tool tip motion is tracked using OptiTrack with an attached marker. Executed trajectory fitting results for \textbf{(i)} line and \textbf{(ii)} circle motions. Each trajectory is repeated three times.}
	\label{fig:benchtop} 
\end{figure}

\textbf{Command-execution tracking:} The second component evaluates how accurately the surgical instrument tool tip follows the commanded input. Tracking performance is quantified by commanding the tool to execute repeatable planar trajectories while maintaining the RCM constraint. The executed tool-tip trajectory is tracked using an OptiTrack motion-capture system (NaturalPoint) by attaching a reflective marker to the tool tip, as shown in Fig. \ref{fig:benchtop} D(i).

Since aligning the OptiTrack base-station coordinate frame with the robot base frame would introduce additional physical measurement errors, direct trajectory comparison in Cartesian space becomes unreliable. Instead, precision is measured using RMS geometric residuals relative to fitted motion primitives. The fitted trajectories are obtained via Principal Component Analysis (PCA): perpendicular deviations are reported for lines, while radial deviation and out-of-plane deviation are reported for circles. The detailed derivation is illustrated in the Method section. The predefined primitives include 100 mm straight lines and planar circles with a diameter of 80 mm. The recorded qualitative results are shown in Fig. \ref{fig:benchtop}D(ii, iii).

Across three trials, straight-line motions achieve $1.30 \pm 0.03$ mm RMS orthogonal deviation over a $\sim$94 mm length, while circular motions exhibit $10.40 \pm 1.32$ mm RMS radial deviation from the commanded 80 mm diameter with low out-of-plane error ($1.23 \pm 0.14$ mm RMS). These results show that humanoid teleoperation can achieve millimeter-scale line tracking, but curved motions remain limited by in-plane accuracy. In contrast, clinically teleoperated surgical robots report positional accuracy on the order of $\sim$1 mm under calibrated conditions \cite{ferguson2020comparing}. From the experiments, this gap primarily arises from inaccuracies in the formulated kinematic chain for wristed laparoscopic instrument control. As shown in Fig. \ref{fig:system}(B, C) and Fig. \ref{fig:IK}(A), the tool-tip pose of the manual instrument, observed in the endoscopic view, is used as the control input and is fed into the extended inverse kinematics solver to compute the corresponding robot wrist pose.
This extended kinematic chain incorporates the instrument’s geometric parameters, such as shaft lengths and link offsets. However, since these commercial tools do not provide publicly available manufacturing specifications, all geometric parameters must be manually measured, introducing discrepancies between the modeled kinematic chain and the ground truth.
Furthermore, RCM control is highly sensitive to trocar localization accuracy, particularly in terms of orientation. Due to the lever-like coupling between the robot wrist and the tool tip, even small errors in vision-based ArUco marker detection can be amplified along the kinematic chain. Together, these factors contribute to increased command-execution errors, particularly when compared to integrated, purpose-built surgical robotic systems.


\pagebreak

\subsection{Dry lab evaluation of the framework}

To evaluate the performance and robustness of the proposed humanoid framework relative to existing surgical platforms, we conduct benchmarking user studies with participants spanning multiple surgical skill levels. Tasks include bi-manual rubber ring replacement and Foundations of Laparoscopic Surgery (FLS) peg transfer with quantifiable performance metrics. Participants perform each task across three platforms: humanoid, surgical robot, and manual operation. For humanoid and manual conditions, laparoscopic trainer boxes enforce RCM constraints for manual wristed instruments (Fig.~\ref{fig:user study}A(i)). In dry-lab experiments, the da Vinci Research Kit (dVRK) and da Vinci Xi systems are used as comparative platforms for the o-ring transfer and FLS peg transfer tasks, respectively.

\subsubsection{User study procedure}
Each participant follows a structured and standardized procedure for performance evaluation and data collection, as outlined below.

\textbf{Pre-experiment:} Before the official trials, all participants complete a standardized familiarization session for each platform. For the robotic systems, this training lasts approximately five minutes each and covers fundamental operations, including handle-based motion control, foot pedal usage, vision system interaction, and safety considerations. For the manual condition, participants spend an equivalent amount of time practicing basic manipulation with wristed laparoscopic tools.

\textbf{Trials \& recording:} After platform familiarization, participants conduct the official trials, and both task completion time and operation performance are recorded for evaluation. The timer starts when the tools first make contact with the initial object. 

\textbf{Platform rotation:} Participants repeat the pre-experiment training and trial procedures on all three platforms (humanoid, surgical robot, and manual) in a randomized order to reduce potential performance degradation due to prolonged operation and fatigue.



\textbf{Post-experiment survey:} Upon completing all required trials, participants fill out a post-experiment questionnaire to assess their experience with each platform. A Likert-scale survey was used to capture user experience, focusing on mental demand, physical demand, motion accuracy, feedback quality, and overall performance for specific workflow components. These pilot data provide targeted insights into the humanoid platform’s usability and task-specific performance. 



\subsubsection{User study task: ring transfer}
Following a similar pattern to \cite{richter2019motion}, the task is a standardized ring transfer procedure. As illustrated in Fig.~\ref{fig:user study}A(ii), the experimental setup consists of a training board with four pegs arranged in a square configuration, each separated by 40 mm. Two rubber o-rings are placed on the pegs as the task objects, with straws positioned beneath them to provide sufficient grasping clearance from the board surface. 
Each trial includes the following steps: 1. Lift a rubber o-ring from either the front (back), left, or right peg using the corresponding arm. 2. Pass the o-ring to the other arm. 3. Place the o-ring on the opposite front (back) peg. 4.  Repeat the same procedure for the back (front) pair of pegs. Each participant completed 6 trials in total.

A total of 18 participants took part in this task, including 6 surgeons and 12 novices who had not received any medical training.  dVRK in the engineering lab is used as the comparative surgical robot platform for this task. 

 In this task, no hard failure condition is defined; instead, performance is evaluated using weighted error scores and total task completion time. Weighted penalties are assigned to common operation errors to reflect severity: [failed ring pick-up: 2], [stretch ring on pegs: 2], [stretch ring during hand-off: 4], [drop ring (outside of pegs): 5], [collision (pegs, ground, tools): 3], and [straw displacement: 3].


The performance statistics across platforms are summarized in Table~\ref{tab:oring}, which reports the per-trial operation times, weighted error scores, and error occurrence counts, along with paired comparison statistics including mean differences, 95\% confidence intervals, effect sizes, and statistical test results. The corresponding boxplot distributions are shown in Fig.~\ref{fig:user study} B(i, ii). Overall, manual operation achieves moderate completion time ($64.22 \pm 31.82$ s) but exhibits the highest weighted error and variability ($7.03 \pm 4.88$). In contrast, both the humanoid platform ($4.53 \pm 3.14$) and the dVRK ($4.59 \pm 2.53$) achieve lower weighted errors, with comparable performance between the two, while the dVRK provides the fastest completion time ($43.01 \pm 15.25$ s).

In the novice group, the dVRK achieves both the fastest completion time and the lowest error, while the humanoid platform demonstrates intermediate accuracy but requires the longest time. In the surgeon group, performance is generally more accurate and consistent across all platforms; surgeons achieve their lowest error on the humanoid platform while remaining fastest on the dVRK, following a similar timing trend as observed in novices. 
Paired comparisons show significant reductions in weighted error for both the humanoid platform and the dVRK compared to manual operation. Completion times differ substantially across platforms: the dVRK is faster than both manual operation and the humanoid platform. 

Overall, these results highlight the performance improvements of the humanoid platform over manual operation while also underscoring the need for further development to achieve more responsive and precise control.

\subsubsection{User study task: FLS peg transfer}

\begin{figure}[!htbp] 
	\centering
	\includegraphics[width=0.85\textwidth]{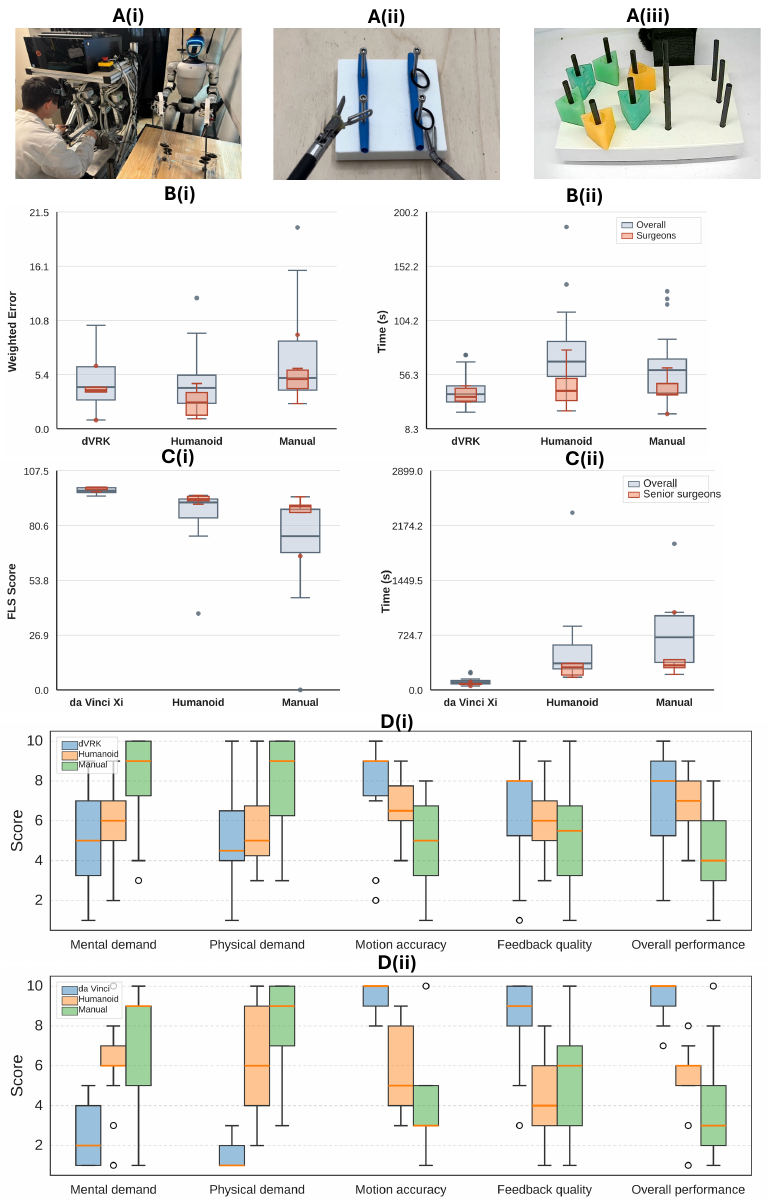} 

	\caption{\textbf{Operation performance, and post-study survey results are reported for user study tasks.} \textbf{(A)(i-iii)} User study conditions. \textbf{(B)} O-ring replacement performance results. Dots indicate individual trials outside $1.5 \times$ the interquartile range (IQR). \textbf{(C)} FLS peg transfer performance results.  \textbf{(D)} Post study survey results of the ring replacement \textbf{(i)} and FLS peg transfer \textbf{(ii)}.}
	\label{fig:user study} 
\end{figure}

To further validate the proposed framework on participants with stronger medical backgrounds, the classic Fundamentals of Laparoscopic Surgery (FLS) peg transfer task is included in the user study for its more diverse wrist motions and well-defined quantitative evaluation protocol. 13 participants with professional medical training took part, including 8 junior surgeons and 5 senior surgeons.
As illustrated in Fig.~\ref{fig:user study}A(iii), the task consists of bimanually transferring all six pegs from one side of the training board to the other and repeating it in the opposite direction. 

To comprehensively evaluate efficiency and safety, task performance is quantified using completion time, operational errors, and the FLS score. Operational errors are defined as peg drops, unintentional collisions, and failed peg pickups. The occurrences of each error type are reported in Table \ref{tab:fls}(B).
The FLS peg transfer task conventionally has a 300~s time limit, and both completion time and errors are considered in the performance evaluation \cite{fls_guidelines_2014}. A commonly used scoring formulation is  $300 - t - 10e$, where $t$ is the completion time (s) and $e$ is the number of errors.
However, completion times for humanoid and manual operation frequently exceed the 300~s limit (Table~\ref{tab:fls}), applying this metric would produce negative or otherwise unbalanced scores.  To provide a single interpretable measure that reflects both speed and reliability, the balanced FLS score is computed by min--max normalizing time and error across all valid trials and combining them with equal weights, then rescaling to $[0,100]$, where higher values indicate faster completion with fewer drops. The score can be calculated as
\begin{equation}
t'=\frac{t-t_{\min}}{t_{\max}-t_{\min}},\qquad
e'=\frac{e-e_{\min}}{e_{\max}-e_{\min}}
\end{equation}

\begin{equation}
\mathrm{S}
=100\left(1-\left(w\,t' + (1-w)\,e'\right)\right),
\qquad w=0.5
\end{equation}
where $t$ is the completion time (s), $\mathrm{S}$ is the balanced FLS score, $e$ is the number of drops in the trial, and $t_{\min}, t_{\max}, e_{\min}, e_{\max}$ are the minimum and maximum values of completion time and peg drops across all participants. $w = 0.5$ is the weighting factor that balances the relative contributions of time and drops.

The results are reported in Table~\ref{tab:fls} and Fig.~\ref{fig:user study}C(i, ii). Overall, a clear performance hierarchy is observed across platforms. The humanoid platform demonstrates reduced completion time and lower total errors relative to manual operation, resulting in an improved FLS score ($85.39 \pm 16.25$ vs. $70.47 \pm 27.53$). However, it remains notably slower and less accurate than the da~Vinci~Xi system, which achieves the best overall performance with the shortest completion time ($118.2 \pm 56.7$~s), lowest error ($1.33 \pm 1.23$), and highest FLS score ($97.67 \pm 1.54$). The trend is consistent across experience levels, with senior surgeons achieving better overall performance across all three platforms, likely due to greater familiarity and experience with both robotic and manual laparoscopic operations. Paired comparisons confirm that the da~Vinci~Xi significantly outperforms both the humanoid and manual conditions across all metrics, while the humanoid platform also shows significant improvements over manual operation. In addition, the five participating senior surgeons completed a task-oriented workload assessment questionnaire across platforms, as summarized in Table~\ref{tab:workload}. The subjective feedback aligns with the quantitative performance metrics.

Overall, these results reinforce a consistent performance separation, with the da~Vinci Xi platform achieving the highest performance, manual operation the lowest, and the humanoid platform falling in between. Compared to the first task, The larger performance gap between the da~Vinci Xi and humanoid platforms is likely attributable to the greater commercial maturity of the da~Vinci~Xi system compared with the research-oriented dVRK platform used in the first task.

The results from both tasks collectively demonstrate the current performance of the humanoid platform for laparoscopic-style manipulation. In the O-ring replacement task, novice error rates on the humanoid are comparable to the dVRK and substantially lower than manual operation, while expert surgeons achieve their lowest errors on the humanoid, indicating strong support for precise manipulation with sufficient expertise. In the more challenging FLS peg transfer task, the humanoid shows intermediate performance, outperforming manual operation in both time and error while remaining behind the more mature da Vinci Xi system.

Post-participation questionnaire results are shown in Fig.~\ref{fig:user study}(D)(i, ii), summarizing participants' system assessment across the three platforms. Overall, participants reported improved performance with the humanoid platform compared with manual operation. Key limitations include reduced responsiveness in control feedback introduced by latency, a less intuitive control interface compared with the da Vinci Surgical System, and constrained reachability resulting from the humanoid’s compact form factor. This feedback aligns with the quantitative benchtop analysis discussed in previous sections, where higher latency and motion error require greater operator skill to close the control loop. Additionally, the Unitree G1 humanoid has an arm span of approximately $450$~mm, compared with $1.6$--$1.8$~m for an adult human \cite{fryar2021anthropometric}, a constraint that becomes particularly evident in dry-lab settings designed for manual skill training.

\pagebreak

\pagebreak

\begin{table*}[t]
\centering
\caption{\textbf{O-ring transfer performance across platforms and expertise levels.(n = 18)}
(A) Performance summary. (B) Error type occurrences.  (C) Paired comparison statistics.  
In this task, participants include 12 novices and 6 surgeons.}
\label{tab:oring}
\footnotesize
\setlength{\tabcolsep}{3pt}
\renewcommand{\arraystretch}{0.85}

\begin{minipage}{\textwidth}
\centering
\begin{tabular*}{\linewidth}{@{\extracolsep{\fill}}@{}llcc@{}}
\toprule
\multicolumn{4}{@{}l@{}}{\textbf{(A) O-ring transfer: weighted error and time}} \\
\midrule
\textbf{Group} & \textbf{Platform} & \textbf{Weighted Error $\downarrow$} & \textbf{Time (s) $\downarrow$} \\
\midrule
\multirow{3}{*}{Overall}
& Manual   & $7.03 \pm 4.88$ & $64.22 \pm 31.82$ \\
& Humanoid & $4.53 \pm 3.14$ & $74.67 \pm 39.22$ \\
& dVRK     & $4.59 \pm 2.53$ & $43.01 \pm 15.25$ \\
\midrule
\multirow{3}{*}{Novices}
& Manual   & $7.91 \pm 5.64$ & $75.43 \pm 32.66$ \\
& Humanoid & $5.49 \pm 3.35$ & $89.23 \pm 38.91$ \\
& dVRK     & $5.00 \pm 2.83$ & $45.37 \pm 17.93$ \\
\midrule
\multirow{3}{*}{Surgeons}
& Manual   & $5.26 \pm 2.34$ & $41.79 \pm 13.79$ \\
& Humanoid & $2.61 \pm 1.44$ & $45.54 \pm 19.25$ \\
& dVRK     & $3.77 \pm 1.72$ & $38.30 \pm 6.57$ \\
\bottomrule
\end{tabular*}
\end{minipage}

\vspace{3pt}
\begin{minipage}{\textwidth}
\centering
\begin{tabular*}{\linewidth}{@{\extracolsep{\fill}}@{}lcccccc@{}}
\toprule
\multicolumn{7}{@{}l@{}}{\textbf{(B) Error type occurrences}} \\
\midrule
Platform & \makecell{Failed\\pick} &
\makecell{Stretch\\(peg)} &
\makecell{Stretch\\(handoff)} &
Drop &
Collision &
\makecell{Straw\\disp.} \\
\midrule
Manual   & $6.11\pm4.30$ & $5.17\pm4.95$ & $0.83\pm1.04$ & $0.83\pm1.10$ & $7.17\pm5.70$ & $0.28\pm0.46$ \\
Humanoid & $5.44\pm4.63$ & $2.39\pm2.59$ & $0.39\pm0.61$ & $1.11\pm1.23$ & $3.33\pm3.24$ & $0.33\pm0.77$ \\
dVRK     & $3.17\pm2.01$ & $0.78\pm0.88$ & $0.22\pm0.73$ & $0.28\pm0.57$ & $7.44\pm4.59$ & $0.67\pm0.69$ \\
\bottomrule
\end{tabular*}
\end{minipage}

\vspace{3pt}
\begin{minipage}{\textwidth}
\centering
\begin{tabular*}{\linewidth}{@{\extracolsep{\fill}}@{}l l c c c c@{}}
\toprule
\multicolumn{6}{@{}l@{}}{\textbf{(C) Paired statistics (overall)}} \\
\midrule
Metric & Comparison & $\Delta$ [95\% CI] & $p$ & $d_z$ & Test \\
\midrule
Weighted error & Humanoid -- Manual & $-2.50$ [$-4.42$, $-0.58$] & $0.014$ & $0.65$ & t \\
Weighted error  & dVRK -- Manual     & $-2.44$ [$-4.76$, $-0.12$] & $0.040$ & $0.52$ & t \\
Weighted error  & Humanoid -- dVRK   & $-0.06$ [$-1.35$, $1.23$]  & $0.922$ & $0.02$ & t \\
\midrule
Time  & Humanoid -- Manual & $10.45$ [$-5.86$, $26.75$] & $0.194$ & $0.32$ & t \\
Time  & dVRK -- Manual     & $-21.21$ [$-35.37$, $-7.04$] & $0.004$ & $0.74$ & w \\
Time  & Humanoid -- dVRK   & $31.65$ [$17.22$, $46.09$] & $2.41e{-}04$ & $1.09$ & t \\
\bottomrule

\end{tabular*}
\end{minipage}

\vspace{3pt}
\begin{minipage}{\textwidth}
\centering
\begin{tabular*}{\linewidth}{@{\extracolsep{\fill}}@{}l l c c c c@{}}
\toprule
\multicolumn{6}{@{}l@{}}{\textbf{(C2) Paired statistics (Novices)}} \\
\midrule
Metric & Comparison & $\Delta$ [95\% CI] & $p$ & $d_z$ & Test \\
\midrule
Weighted error  & Humanoid -- Manual & $-2.42$ [$-5.30$, $0.45$] & $0.091$ & $0.54$ & t \\
Weighted error  & dVRK -- Manual     & $-2.92$ [$-6.44$, $0.61$] & $0.096$ & $0.53$ & t \\
Weighted error  & Humanoid -- dVRK   & $0.49$ [$-1.38$, $2.36$]  & $0.575$ & $0.17$ & t \\
\midrule
Time  & Humanoid -- Manual & $13.80$ [$-9.56$, $37.16$] & $0.220$ & $0.38$ & t \\
Time  & dVRK -- Manual     & $-30.06$ [$-49.35$, $-10.77$] & $0.006$ & $0.99$ & t \\
Time  & Humanoid -- dVRK   & $43.86$ [$27.87$, $59.85$] & $4.88e{-}04$ & $1.74$ & w \\
\bottomrule
\end{tabular*}
\end{minipage}

\vspace{3pt}

\begin{minipage}{\textwidth}
\centering
\begin{tabular*}{\linewidth}{@{\extracolsep{\fill}}@{}l l c c c c@{}}
\toprule
\multicolumn{6}{@{}l@{}}{\textbf{(C3) Paired statistics (Surgeons)}} \\
\midrule
Metric & Comparison & $\Delta$ [95\% CI] & $p$ & $d_z$ & Test \\
\midrule
Weighted error  & Humanoid -- Manual & $-2.65$ [$-5.08$, $-0.22$] & $0.038$ & $1.15$ & t \\
Weighted error & dVRK -- Manual     & $-1.49$ [$-3.75$, $0.78$]  & $0.094$ & $0.69$ & w \\
Weighted error  & Humanoid -- dVRK   & $-1.16$ [$-2.51$, $0.18$]  & $0.076$ & $0.91$ & t \\
\midrule
Time  & Humanoid -- Manual & $3.74$ [$-21.98$, $29.46$] & $0.724$ & $0.15$ & t \\
Time  & dVRK -- Manual     & $-3.49$ [$-16.98$, $9.99$] & $0.535$ & $0.27$ & t \\
Time  & Humanoid -- dVRK   & $7.23$ [$-13.68$, $28.15$] & $0.688$ & $0.36$ & w \\
\bottomrule
\end{tabular*}
\end{minipage}

\vspace{2pt}
{\scriptsize
\textit{Notes:} $\Delta$ is mean paired difference with 95\% CI. $d_z$ is Cohen’s $d$. Test: paired $t$-test (t) or Wilcoxon (w).
}
\end{table*}

\clearpage

\begin{table*}[t]
\centering
\caption{\textbf{FLS peg transfer performance across platforms and expertise levels (n = 13).}
(A) Performance summary. (B) Error type occurrences report. (C) Paired comparison statistics across expertise levels. In this task, participants include 8 junior surgeons and 5 senior surgeons}
\label{tab:fls}
\vspace{-5pt}
\footnotesize

\begin{minipage}{\textwidth}\centering
\begin{tabular*}{\linewidth}{@{\extracolsep{\fill}}@{}llccc@{}}
\toprule
\multicolumn{5}{@{}l@{}}{\textbf{(A) FLS peg transfer: performance summary (mean $\pm$ std)}} \\
\midrule
\textbf{Group} & \textbf{Platform} & \textbf{Time (s) $\downarrow$} & \textbf{Total Error $\downarrow$} & \makecell{\textbf{FLS Score}\textbf{(0--100)} $\uparrow$} \\
\midrule
\multirow{3}{*}{Overall}
& Manual   & $877.8 \pm 774.3$  & $17.45 \pm 16.80$ & $70.47 \pm 27.53$ \\
& Humanoid & $560.1 \pm 595.5$  & $6.25 \pm 6.36$   & $85.39 \pm 16.25$ \\
& da Vinci Xi & $118.2 \pm 56.7$   & $1.33 \pm 1.23$   & $97.67 \pm 1.54$ \\
\midrule
\multirow{3}{*}{Junior surgeons}
& Manual   & $1234.2 \pm 881.4$ & $24.83 \pm 19.63$ & $57.86 \pm 31.47$ \\
& Humanoid & $765.1 \pm 726.3$  & $8.71 \pm 7.52$   & $79.54 \pm 19.67$ \\
& da~Vinci~Xi & $146.7 \pm 58.1$   & $1.57 \pm 1.27$   & $96.94 \pm 1.47$ \\
\midrule
\multirow{3}{*}{Senior surgeons}
& Manual   & $450.2 \pm 329.5$ & $8.60 \pm 6.62$ & $85.61 \pm 11.49$ \\
& Humanoid & $273.0 \pm 86.6$  & $2.80 \pm 0.84$ & $93.57 \pm 1.73$ \\
& da Vinci Xi & $78.4 \pm 19.6$   & $1.00 \pm 1.22$ & $98.69 \pm 1.04$ \\
\bottomrule
\end{tabular*}
\end{minipage}

\begin{minipage}{\textwidth}\centering
\begin{tabular*}{\linewidth}{@{\extracolsep{\fill}}@{}lccc@{}}
\toprule
\multicolumn{4}{@{}l@{}}{\textbf{(B) Error type occurrences}} \\
\midrule
Platform & Drops & Collision & Failed Pick Up \\
\midrule
Manual   & $3.82 \pm 3.76$ & $8.27 \pm 9.12$ & $5.36 \pm 5.61$ \\
Humanoid & $2.17 \pm 2.76$ & $2.08 \pm 1.98$ & $2.42 \pm 2.39$ \\
da Vinci Xi & $0.17 \pm 0.39$ & $1.00 \pm 1.04$ & $0.17 \pm 0.39$ \\
\bottomrule
\end{tabular*}
\end{minipage}


\begin{minipage}{\textwidth}\centering
\begin{tabular*}{\linewidth}{@{\extracolsep{\fill}}@{}l l c c c c@{}}
\toprule
\multicolumn{6}{@{}l@{}}{\textbf{(C) Paired statistics (overall)}} \\
\midrule
Metric & Comparison & $\Delta$ [95\% CI] & $p$ & $d_z$ & Test \\
\midrule
Time (s) & da Vinci Xi -- Humanoid & $-467.27$ [$-850.44$, $-84.10$]   & $9.77e{-}04$ & $0.82$ & w \\
Time (s) & da Vinci Xi-- Manual   & $-757.73$ [$-1242.53$, $-272.93$] & $9.77e{-}04$ & $1.05$ & w \\
Time (s) & Humanoid -- Manual   & $-290.45$ [$-526.92$, $-53.98$]   & $0.021$      & $0.83$ & t \\
\midrule
Error & da Vinci Xi -- Humanoid & $-5.18$ [$-9.67$, $-0.70$]   & $0.004$      & $0.78$ & w \\
Error & da Vinci Xi -- Manual   & $-16.09$ [$-27.44$, $-4.74$] & $9.77e{-}04$ & $0.95$ & w \\
Error & Humanoid  -- Manual   & $-10.91$ [$-18.03$, $-3.78$] & $9.77e{-}04$ & $1.03$ & w \\
\midrule
\makecell[l]{FLS Score} & da Vinci Xi -- Humanoid & $12.97$ [$2.22$, $23.72$] & $9.77e{-}04$ & $0.81$ & w \\
\makecell[l]{FLS Score} & da Vinci Xi -- Manual   & $27.14$ [$9.21$, $45.06$] & $9.77e{-}04$ & $1.02$ & w \\
\makecell[l]{FLS Score} & Humanoid -- Manual   & $14.17$ [$5.77$, $22.56$] & $0.004$      & $1.13$ & t \\
\bottomrule
\end{tabular*}
\end{minipage}

\begin{minipage}{\textwidth}\centering
\begin{tabular*}{\linewidth}{@{\extracolsep{\fill}}@{}l l c c c c@{}}
\toprule
\multicolumn{6}{@{}l@{}}{\textbf{(C2) Paired statistics (junior surgeons)}} \\
\midrule
Metric & Comparison & $\Delta$ [95\% CI] & $p$ & $d_z$ & Test \\
\midrule
Time & da~Vinci Xi -- Humanoid & $-694.50$ [$-1441.64$, $52.64$] & $0.031$ & $0.98$ & w \\
Time & da~Vinci Xi -- Manual   & $-1079.33$ [$-1947.57$, $-211.10$] & $0.024$ & $1.30$ & t \\
Time & Humanoid -- Manual   & $-384.83$ [$-790.54$, $20.87$] & $0.059$ & $1.00$ & t \\
\midrule
Error & da~Vinci Xi -- Humanoid & $-8.00$ [$-16.55$, $0.55$] & $0.061$ & $0.98$ & t \\
Error & da~Vinci Xi -- Manual   & $-23.17$ [$-44.07$, $-2.27$] & $0.031$ & $1.16$ & w \\
Error & Humanoid -- Manual   & $-15.17$ [$-27.99$, $-2.34$] & $0.031$ & $1.24$ & w \\
\midrule
\makecell[l]{Balanced FLS} & da~Vinci Xi -- Humanoid & $19.52$ [$-1.30$, $40.33$] & $0.031$ & $0.98$ & w \\
\makecell[l]{Balanced FLS} & da~Vinci Xi -- Manual   & $38.85$ [$6.51$, $71.19$] & $0.027$ & $1.26$ & t \\
\makecell[l]{Balanced FLS} & Humanoid -- Manual   & $19.33$ [$6.21$, $32.46$] & $0.013$ & $1.55$ & t \\
\bottomrule
\end{tabular*}
\end{minipage}

\begin{minipage}{\textwidth}\centering
\begin{tabular*}{\linewidth}{@{\extracolsep{\fill}}@{}l l c c c c@{}}
\toprule
\multicolumn{6}{@{}l@{}}{\textbf{(C3) Paired statistics (senior surgeons)}} \\
\midrule
Metric & Comparison & $\Delta$ [95\% CI] & $p$ & $d_z$ & Test \\
\midrule
Time & da Vinci Xi -- Humanoid & $-194.60$ [$-314.32$, $-74.88$] & $0.062$ & $2.02$ & w \\
Time & da Vinci Xi -- Manual   & $-371.80$ [$-774.06$, $30.46$]  & $0.062$ & $1.15$ & w \\
Time & Humanoid -- Manual   & $-177.20$ [$-556.94$, $202.54$] & $0.265$ & $0.58$ & t \\
\midrule
Error & da Vinci Xi -- Humanoid & $-1.80$ [$-3.64$, $0.04$] & $0.053$ & $1.21$ & t \\
Error & da Vinci Xi -- Manual   & $-7.60$ [$-16.58$, $1.38$] & $0.079$ & $1.05$ & t \\
Error & Humanoid -- Manual   & $-5.80$ [$-13.11$, $1.51$] & $0.092$ & $0.98$ & t \\
\midrule
\makecell[l]{Balanced FLS} & da Vinci Xi -- Humanoid & $5.12$ [$1.82$, $8.42$] & $0.013$ & $1.93$ & t \\
\makecell[l]{Balanced FLS} & da Vinci Xi -- Manual   & $13.08$ [$-1.60$, $27.76$] & $0.062$ & $1.11$ & w \\
\makecell[l]{Balanced FLS} & Humanoid -- Manual   & $7.96$ [$-4.86$, $20.79$] & $0.160$ & $0.77$ & t \\
\bottomrule
\end{tabular*}
\end{minipage}

\vspace{4pt}
{\scriptsize
\textit{Notes:} $\Delta$ is the mean paired difference with 95\% CI.
$d_z$ is Cohen’s $d$ for paired samples. Test: paired $t$-test (t) or Wilcoxon signed-rank (w).
}
\end{table*}

\clearpage

\begin{table}[t]
\caption{\textbf{Task workload assessment by senior surgeons.} Subjective workload assessment (mean $\pm$ standard deviation) across platforms based on post-study questionnaires.}
\label{tab:workload}

\centering

\begin{tabular*}{\linewidth}{@{\extracolsep{\fill}}@{}l c c c@{}}
\toprule
Metric & da~Vinci~Xi & Humanoid & Manual \\
\midrule
Mental demand $\downarrow$  & $2.00 \pm 1.22$ & $5.40 \pm 1.34$ & $8.80 \pm 0.84$ \\
Physical demand $\downarrow$ & $1.60 \pm 0.89$ & $4.40 \pm 2.61$ & $8.20 \pm 0.84$ \\
Temporal demand $\downarrow$ & $1.60 \pm 0.89$ & $4.00 \pm 1.58$ & $7.40 \pm 3.13$ \\
Overall performance $\uparrow$ & $9.20 \pm 1.30$ & $6.00 \pm 1.22$ & $3.40 \pm 1.14$ \\
Effort $\downarrow$     & $2.20 \pm 1.30$ & $5.60 \pm 1.82$ & $8.80 \pm 1.10$ \\
Frustration $\downarrow$   & $1.40 \pm 0.89$ & $4.20 \pm 0.84$ & $8.60 \pm 1.34$ \\
\bottomrule
\end{tabular*}
\end{table}

\clearpage

\subsection{Live porcine surgery}

\begin{figure}[!htbp] 
	\centering
	\includegraphics[width=1\textwidth]{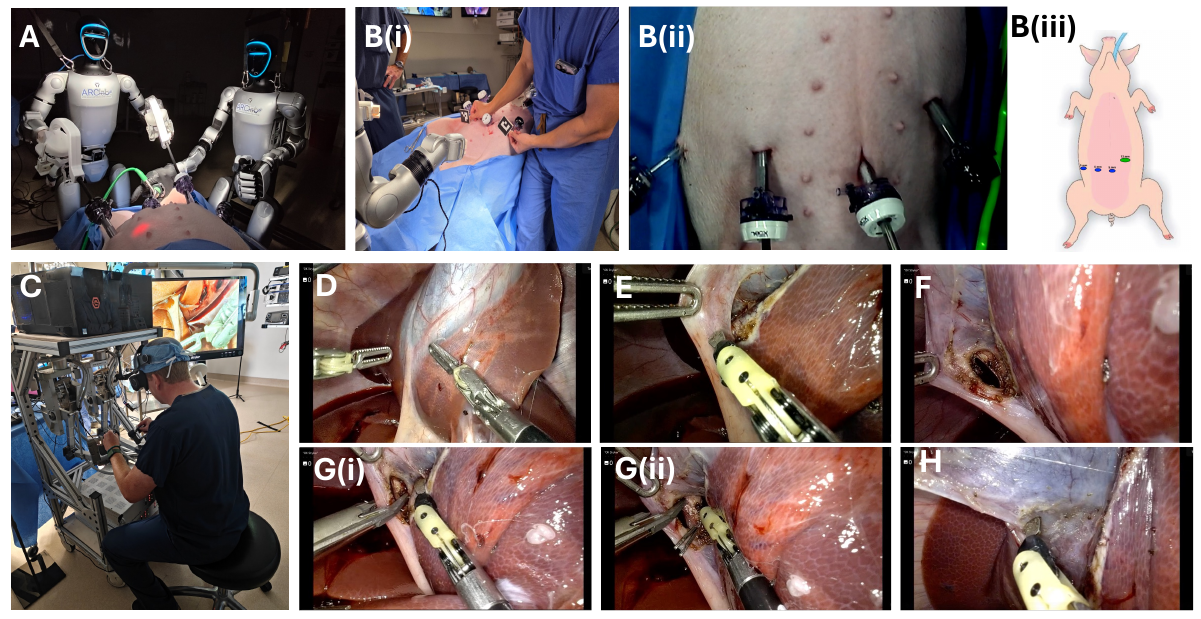} 

	\caption{\textbf{Humanoids performing live porcine laparoscopic cholecystectomies.} (\textbf{A}) Two humanoid robots performing bimanual tasks at the bedside. The surgeon humanoid operates laparoscopic instruments, while the assistant robot provides camera control and tissue retraction. (\textbf{B(i-iii)}) Initial system deployment. \textbf{(i)}External RCM calibration for port set-up using ArUco markers. \textbf{(ii, iii)} Port placement setup. (\textbf{C}) A senior surgeon operating a surgeon humanoid from the control console.  (\textbf{D-H}) Laparoscopic footage demonstrating (\textbf{D})  initial retraction, (\textbf{E}) initial dissection, (\textbf{D})  critical view of safety, (\textbf{G(i, ii)}) laparoscopic clipping of the cystic duct, and (\textbf{H}) gallbladder dissection off the liver bed.}
	\label{fig:surgery} 
\end{figure}

Two live porcine laparoscopic cholecystectomies were performed using the humanoid robotic surgical system in our institution’s surgical simulation center, which provides fully equipped operating rooms with standard laparoscopic, robotic, and endoscopic infrastructure. All procedures were conducted with institutional animal care approval (UCSD IACUC \#S08027). A licensed veterinarian administered and monitored the animals’ general anesthesia, while simulation center staff assisted with sterile preparation and draping of the animal before incision. The procedure is illustrated in Fig.~\ref{fig:surgery}.

Abdominal access and port placement were performed by attending surgeons and clinical fellows using a conventional robotic-assisted laparoscopic cholecystectomy configuration with four trocars (three 5 mm, one 12 mm). As shown in Fig.~\ref{fig:surgery}\textbf{B(i)}, the ports are localized before the operation by attaching calibration boards with ArUco markers, which are detected by the humanoid’s head camera. After port placement, the humanoid robot was positioned at the bedside and aligned with the operative field. In this study, a harness was used as a precautionary safeguard, which is not required for system functionality. Deployment required close coordination between surgical and engineering teams, with iterative adjustments to optimize robot positioning, instrument alignment, and range of motion relative to the trocars.  

Procedures followed a standard laparoscopic workflow, with one console surgeon and one bedside assistant. To ensure consistency, a single senior surgeon performed all procedures from the robotic console using the humanoid platform, while bedside assistance was provided by either another senior surgeon or a clinical fellow experienced in laparoscopic surgery. The assistant managed camera control, tissue retraction, exposure, and bedside adjustments such as camera cleaning and instrument adjustments. Instrument adjustments and camera cleaning followed standard laparoscopic techniques and typically required less than 1 minute; these events were recorded and analyzed as micro-pauses in the workflow analysis. During a brief interval in the first case, a second humanoid robot was used for camera holding and retraction; however, most bedside assistance in both cases was provided by a human assistant.

Due to the animal's respiration and slight robot base drift during the operation, the actual RCM positions were dynamic. A temporal filter on the estimated RCM positions was adopted to prevent unsafe disturbances. Procedures were paused for recalibration in cases of large RCM drift, degraded visualization, unexpected instrument behavior, or at the surgeon’s discretion. During these pauses, instruments were held in a stable position, and system functionality was verified before resuming the procedure. Instrument exchanges and robot redeployments for recalibration or repositioning typically required at least three minutes and were classified as major pauses in the workflow analysis.


Both procedures followed a standard robotic-assisted laparoscopic cholecystectomy workflow, including gallbladder retraction, dissection of the hepatocystic triangle, identification of the cystic structures, confirmation of the critical view of safety, and mobilization of the gallbladder off the liver bed. Specialized robot-compatible wristed instruments were used for the operation along with standardized laparoscopic ports, cameras, and insufflation devices. Since robot-compatible clip appliers were unavailable, cystic duct clipping and cystic artery ligation were performed laparoscopically before resuming gallbladder mobilization robotically; this was accomplished by removing one robotic arm from the surgical field without having to “undock” the robot.


Quantitative assessment of surgical performance, including the duration of active console time when the surgeon directly manipulated instruments with the humanoid robot, is summarized in  Table \ref{tab:live_porcine_metrics}. Segment-specific analysis of active console times confirmed that tissue dissection, attainment of the critical view of safety, laparoscopic clipping, and gallbladder mobilization off the liver bed were all performed rapidly, highlighting the system’s fidelity and precise instrument control.

\pagebreak
\begin{table}[ht]
\centering
\caption{\textbf{Quantitative metrics of humanoid robotic performance during live porcine cholecystectomies.} Times reflect durations of active console control and segment-specific operative steps. Major pauses ($>3$ min) indicate robot deployments or instrument exchanges; minor pauses ($<1$ min) indicate brief camera or instrument adjustments performed by the bedside assistant. Deployments were defined as discrete episodes of robot recalibration or physical repositioning relative to trocar placement. $\Delta$ denotes the change in each metric between Case 1 and Case 2.}
\label{tab:live_porcine_metrics}
\renewcommand{\arraystretch}{1.15}
\begin{tabularx}{\linewidth}{@{}Xccc@{}}
\toprule
\textbf{Metric} & \textbf{Case 1} & \textbf{Case 2} & \textbf{$\Delta$ (Change)} \\
Active console time (mm:ss) & 56:15 & 31:59 & $-24$:$16$ \\
Robot initiation $\rightarrow$ initiation of tissue dissection (mm:ss) & 4:31 & 2:12 & $-2$:$19$ \\
Initiation of tissue dissection $\rightarrow$ critical view of safety (mm:ss) & 15:42 & 6:19 & $-9$:$23$ \\
Critical view of safety $\rightarrow$ completion of lap clipping (mm:ss) & 4:04 & 3:19 & $-0$:$45$ \\
Gallbladder mobilization off liver bed (mm:ss) & 31:58 & 23:08 & $-8$:$50$ \\
Number of robot deployments (n) & 8 & 4 & $-4$ \\
Number of major pauses ($>3$ min, n) & 2 & 3 & $+1$ \\
Number of minor pauses ($<1$ min, n) & 7 & 4 & $-3$ \\
\bottomrule
\end{tabularx}
\end{table}

\pagebreak

We performed a workflow analysis to identify potential inefficiencies during these procedures. Brief instrument adjustments and camera cleaning events lasting less than 1 minute were classified as micro-pauses. Similar to those routinely performed by human assistants in laparoscopic surgery \cite{nabeel2024assessing, holinka1984diamine, von2016microcomplications}, these reflect normal, expected pauses in the surgical workflow rather than limitations of the humanoid platform. In contrast, major pauses lasting more than three minutes corresponded to full redeployments of the robot for system recalibration or physical repositioning of the robot base or arms relative to the trocars. Unlike routine micro-pauses, these longer interruptions disrupt procedural flow and highlight areas where the technology requires improvement before clinical implementation. Importantly, these redeployments in each case did not affect performance during active console use, and both procedures maintained excellent execution of critical operative tasks (Table \ref{tab:live_porcine_metrics}).

Both cases were completed robotically, demonstrating the initial feasibility of the humanoid robotic system for in vivo use. The first case was completed without major intraoperative complications. In the second case, minor biliary spillage and liver-bed bleeding occurred, which were managed with suction and electrocautery. Neither procedure required conversion to conventional laparoscopy or open surgery. Predefined criteria for conversion included (1) hemodynamic instability of the animal, (2) inability to safely continue the planned surgical task due to inadequate visualization or operative workspace, or (3) technical limitations of the robotic platform preventing safe completion of the procedure.

\subsubsection{User experience and system performance evaluation}
A structured post-procedure survey was distributed to attending surgeons and surgical trainees involved in system setup, bedside assistance, or console operation. Each participant first indicated which workflow components they performed, and survey responses were collected only for those tasks. Consequently, reported scores reflect feedback exclusively from the users who directly performed each component of the procedure, providing a task-specific assessment of usability and workflow performance.

The survey was designed to evaluate four key aspects of system performance:
\begin{enumerate}
    \item \textbf{System effectiveness}: the system’s ability to perform specific tasks.
    \item \textbf{Clinical readiness}: whether the system can be used safely and efficiently in a clinical setting.
    \item \textbf{Ergonomics}: the comfort and usability of the system.
    \item \textbf{Fidelity of motion}: the extent to which intracorporeal instrument motions accurately replicate movements made at the surgeon console.
\end{enumerate}

Survey responses from four participants are summarized in Table~\ref{tab:post_survey}. Quantitative ratings indicated that the system was moderately effective for core operative tasks, with some remaining operational challenges for full clinical deployment. Qualitative feedback contextualized these findings: respondents described the system as capable of performing core operative tasks, with smooth dissection and instrument control when operating within the system’s constraints. At the same time, they emphasized that range-of-motion and strength limitations required frequent repositioning and heightened vigilance from the operating surgeon, even during a relatively localized procedure such as a cholecystectomy. Workflow interruptions related to system recalibration, targeting, and intermittent overheating were observed, though one respondent noted improved performance in the second procedure. Users emphasized that these pauses reflected a major limitation of the current system that would need to be addressed before clinical use in human patients. 
 
Respondents highlighted several strengths of the system, including its compact physical footprint, intuitive console interface, and effective remote center pairing. Users suggested that improving the system’s range of motion, enhancing instrument exchange efficiency, and ensuring compatibility with standard laparoscopic instruments would facilitate future translation for clinical use. These findings provide actionable insights for the iterative system optimization that is needed to transition this system from early feasibility studies to readiness for clinical use.


\pagebreak

\begin{table}[!htbp]
\footnotesize
\centering
   \caption{\textbf{Post-procedure survey results evaluating the humanoid robotic surgical system across setup, bedside assistance, and console operation.} 
    Survey responses ($n=4$) were collected from attending surgeons and surgical trainees involved in system setup (1), bedside assistance (2), or console operation (1). Participants completed only the questions corresponding to the role they performed. Ratings were provided on a Likert scale from 1 (poor/low readiness) to 5 (excellent/high readiness).}
\label{tab:post_survey}

\begin{tabular}{lccc}
\hline
\textbf{Survey Metric} & \textbf{Respondents ($n$)} & \textbf{Mean Effectiveness} & \textbf{Mean Clinical Readiness} \\
\hline
Setup, port placement, docking & 1 & 4 & 3 \\
Ease of bedside tool insertion \& attachment & 1 & 3 & -- \\
Bedside assistance with system & 2 & 3 & 2.5 \\
Ergonomics of bedside assistance & 2 & 3  & -- \\
Surgeon console effectiveness & 1 & 3 & 3 \\
Surgeon console intuitiveness & 1 & 4 & -- \\
Fidelity of motion of surgeon console & 1 & 3 & -- \\
Overall clinical readiness & 4 & -- & 2.5 \\
\hline
\end{tabular}
\end{table}

\clearpage

\section{Discussion}\label{sec12}

In this study, we present a comprehensive evaluation of contemporary humanoid technology for surgical applications and report the first in vivo use of a humanoid robotic surgical system to perform standard laparoscopic cholecystectomies in a porcine model. 

Before in vivo deployment, benchtop experiments and dry-lab user studies provided controlled evaluations of the platform’s technical capabilities and usability. These experiments demonstrated the feasibility of accurately manipulating manual laparoscopic instruments under externally enforced RCM constraints. At the same time, quantitative analyses revealed performance gaps compared with established surgical robotic systems, including control latency, reduced workspace reachability, and sensitivity to calibration drift.

In the in vivo study, both procedures were completed without conversion to conventional laparoscopy or open surgery, demonstrating the system’s capacity for meticulous hepatocystic dissection and gallbladder mobilization under physiologic conditions.
Quantitative assessment of console times confirmed that core operative steps were executed
efficiently and without critical intraoperative errors, indicating that the humanoid platform
facilitates precise, controlled, and reproducible dissection with a level of instrument fidelity
seldom observed in early-stage robotic systems.

Workflow analysis revealed that active console time was efficient, supporting 
effective robotic manipulation of core surgical tasks. Micro-pauses ($<$ minute) were observed
when the bedside assistant performed instrument adjustments and camera cleaning, whereas
major pauses ($>$ 3 minutes) corresponded to workflow interruptions due to robot redeployment or
recalibration. While these interruptions are expected for an early-stage system, they underscore
areas requiring workflow optimization. Surgeon feedback highlighted a distinct gap between
technical feasibility and clinical readiness: constraints in range of motion, force generation, and
the need for frequent recalibration increased both cognitive and operational burden. These
limitations echo early experiences with other robotic platforms, where constrained articulation,
instrument limitations, inconsistent performance, and workflow friction restricted initial clinical
utility \cite{RojasBurbano2024,Kim2002,Rojas2021}. Collectively, these findings delineate specific targets for engineering
improvements necessary before clinical deployment in human patients.

Integrating humanoid robots into sterile operating rooms introduces additional challenges. In this
study, sterility was maintained by placing gloves over the robot’s arms, a strategy that does not
fully replicate sterile workflows required in human surgery, where all components contacting the
operative field must be autoclaved and appropriately draped. Current commercial humanoid
systems lack autoclavable components, and preserving sterility without impairing sensor
function or motion calibration remains a critical barrier. Transparent sterile drapes may offer a
potential solution; however, future work must establish robust protocols for fully sterile
integration while maintaining sensor fidelity.

Once clinically ready, humanoid platforms may help address several limitations inherent to contemporary robotic surgery. High acquisition costs, proprietary consumables, ongoing maintenance contracts, and extensive training requirements have historically confined robotic surgery to well-resourced tertiary centers \cite{burbano2025robot,olawade2025robotic}.
By providing a versatile humanoid platform at a lower cost, these systems may improve accessibility and broaden the availability of robotic-assisted surgical care. Additionally, the humanoid robot’s reduced footprint and positional adaptability may reduce crowding at the operating table, potentially providing bedside assistants greater freedom to perform tasks such as suction, retraction, and instrument exchange. The absence of large bedside consoles or obstructive robotic arms may also help preserve sightlines, which could facilitate team situational awareness, learner engagement, and intraoperative communication.

This study establishes an early, quantitative benchmark for humanoid surgical performance in
vivo. Active console efficiency and precise instrument manipulation were achievable even in
these preliminary deployments, demonstrating feasibility while identifying areas for technical
and workflow refinement. With further development, humanoid systems may provide scalable
alternatives to conventional robotic platforms and expand robotic capabilities beyond traditional
operating rooms. In the future, improved system integration for humanoid manipulation of surgical instruments will be critical for enhancing control accuracy and operational safety. This includes the use of open-source laparoscopic instruments with well characterized geometries, as well as more accurate RCM localization techniques. Future work should also focus on optimizing control strategies to reduce user-perceived latency, improving robustness under noisy or dynamic RCM conditions, and exploring human-scale humanoids to better align with surgical workflows. These findings highlight key technical challenges that must be addressed prior to clinical deployment.

At our institution, close clinician–engineer collaboration has enabled rapid, iterative testing in realistic surgical environments, laying the foundation for continued
refinement. Future investigations will evaluate performance across a broader range of
procedures, extended case durations, and more autonomous tasks to assess the capacity of
humanoid surgical robots to address remaining clinical, economic, and structural gaps in
contemporary robotic surgery.

\pagebreak

\section{Methods}\label{sec11}
\subsection{Humanoid Laparoscopic Teleoperation System}

As shown in Fig.~\ref{fig:system}, the mobile operating console consists of two Master Tool Manipulator (MTM) modules~\cite{kazanzides2014open} that provide the teleoperation interface for two laparoscopic instruments, a foot pedal for controlling instrument in/out-clutching, and a video headset for providing endoscopic views of the scene, and the control workstation that runs teleoperation communication channel, inverse kinematics, and control systems. The vision module combines a dual-1920p stereo-laparoscopic camera with a GOOVIS G3 Max stereo-HD headset, providing clear and low-latency visual feedback for laparoscopic teleoperation. 

On the execution side, the humanoid platform performs bimanual laparoscopic manipulation using commercially available non-robotic wristed instruments by LivsMed (Fig.~\ref{fig:system}(B)). These instruments are designed for human operation and lack active actuation; instead, they couple human wrist motion to wristed end-effector motion through mechanical linkages. In order to achieve similar controllability, a custom mount was developed that interfaces with the humanoid hand while preserving full wrist articulation so that the humanoid was able to achieve all wrist degrees of freedom of the end-effector actively.


Unlike in standard robotic surgery with a daVinci Surgical System, where a remote center of motion is mechanically fixed, surgeons operating laparoscopic instruments must actively maintain a virtual remote center of motion for their laparoscopic tools centered around the trocar ports that interface external and internal cavities of the body. Similarly, the humanoid platform must also create a virtual remote center, and thus instrument control is formulated as an inverse mapping of the extended kinematic chain subject to a remote center of motion (RCM) constraint, as illustrated in Fig.~\ref{fig:system}(C). 
The humanoid wrist pose $P_{ee}$ is regulated such that the instrument shaft pivots about a fixed trocar point, while the instrument tip maps the target pose $P_{tt}$ through coordinated horizontal and vertical motions. 
Since the laparoscopic instruments are non-actuated, jaw opening and closing are achieved via a servo-driven finger interface that engages the instrument's finger holes. 
The grasper angle is mapped from the MTM input to a $0^\circ$--$45^\circ$ range, with real-time streaming of angle commands to ensure precise and responsive grasp control.

The communication between the operating console and the humanoid platform is implemented through the ROS2 network, enabling synchronized bimanual manipulation.

\subsection{Motion precision evaluation}
This section details the mathematical formulation of the motion precision metrics used to evaluate geometric execution accuracy, as summarized in the main text. Let $\{\mathbf{p}_i \in \mathbb{R}^3\}_{i=1}^N$ denote the measured trajectory samples.

\paragraph{Line fitting}

For straight-line motions, a best-fit line is estimated using principal component analysis (PCA). The trajectory centroid is computed as
\begin{equation}
\boldsymbol{\mu} = \frac{1}{N} \sum_{i=1}^N \mathbf{p}_i ,
\end{equation}
and the covariance matrix is given by
\begin{equation}
\mathbf{C} = \frac{1}{N} \sum_{i=1}^N (\mathbf{p}_i - \boldsymbol{\mu})(\mathbf{p}_i - \boldsymbol{\mu})^\top .
\end{equation}
The dominant eigenvector $\mathbf{v}_1$ of $\mathbf{C}$ defines the direction of the best-fit line. The perpendicular deviation of each point from the fitted line is computed as
\begin{equation}
d_i =
\left\| (\mathbf{p}_i - \boldsymbol{\mu}) -
\big((\mathbf{p}_i - \boldsymbol{\mu})^\top \mathbf{v}_1\big)\mathbf{v}_1
\right\| .
\end{equation}
Motion precision for line tracking is reported as the root-mean-square (RMS) of $\{d_i\}$.

\paragraph{Circle fitting}

For circular motions with known radius $R$, the trajectory is first projected onto its best-fit plane obtained via PCA. The plane normal $\mathbf{n}$ is defined as the eigenvector associated with the smallest eigenvalue of $\mathbf{C}$, and an orthonormal basis $(\mathbf{u}, \mathbf{v})$ spanning the plane is constructed from the remaining eigenvectors.

Projected coordinates are computed as
\begin{equation}
(x_i, y_i) = \big((\mathbf{p}_i - \boldsymbol{\mu})^\top \mathbf{u},\;
(\mathbf{p}_i - \boldsymbol{\mu})^\top \mathbf{v}\big).
\end{equation}
The circle center $(c_x, c_y)$ is estimated by solving the least-squares problem
\begin{equation}
(c_x, c_y) = 
\min_{c_x,c_y} \sum_{i=1}^N
\left( \sqrt{(x_i - c_x)^2 + (y_i - c_y)^2} - R \right)^2 .
\end{equation}
Radial deviation is defined as $r_i - R$, where $r_i$ is the distance from $(x_i, y_i)$ to the fitted center. Motion precision for circular trajectories is reported as the RMS radial deviation. Planarity is quantified by the RMS out-of-plane deviation
\begin{equation}
e_{\perp} =
\sqrt{\frac{1}{N} \sum_{i=1}^N \big((\mathbf{p}_i - \boldsymbol{\mu})^\top \mathbf{n}\big)^2 } .
\end{equation}



\subsection{Inverse Mapping Control of Manual Laparoscopic Tools}

\begin{figure}[!htbp] 
	\centering
	\includegraphics[width=1\textwidth]{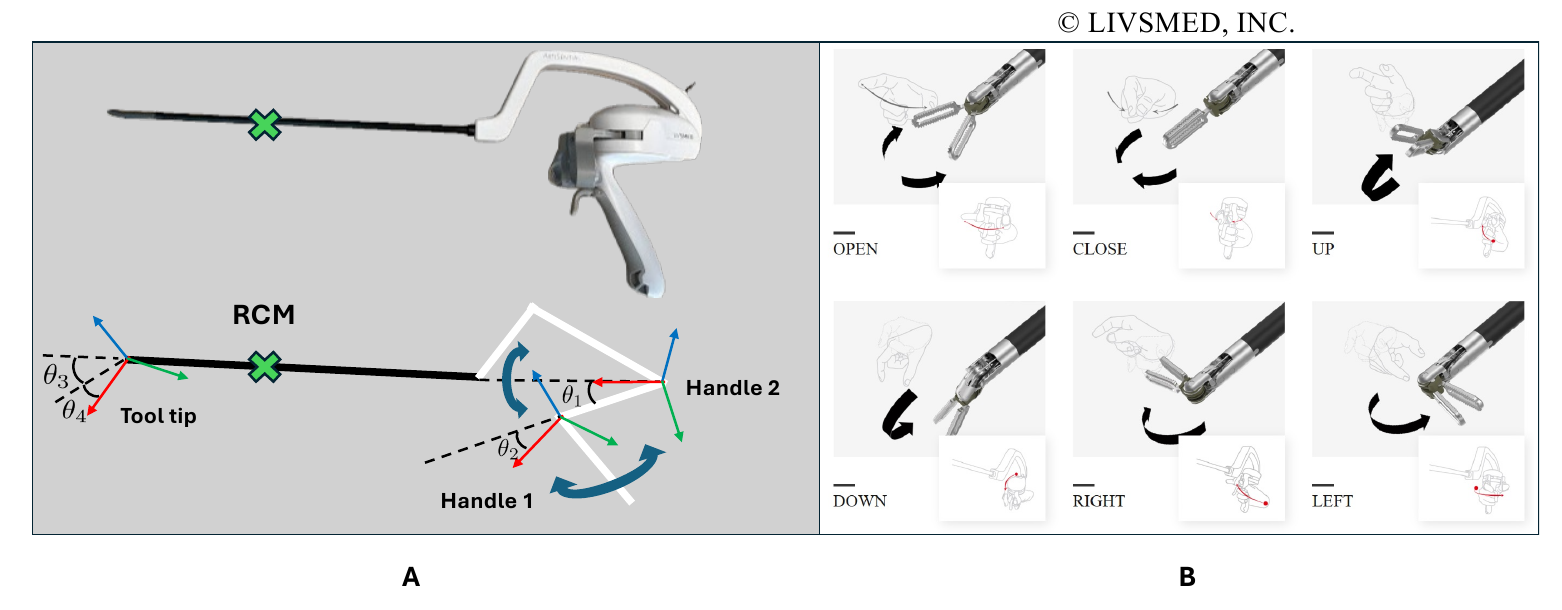} 

	\caption{\textbf{Modeling of the extended kinematic chain.}
		(\textbf{A})The non-actuated instrument orientation is parameterized by two independent orthogonal rotations, $\theta_1$ and $\theta_2$, determined by the handle pose and RCM position. (\textbf{B}) Operation of wristed laparoscopic instruments.}
	\label{fig:IK} 
\end{figure}

In this work, ArtiSential’s bipolar fenestrated forceps are adopted as the commercially available wristed laparoscopic instrument. ArtiSential is FDA 510(k)-cleared and commercialized by LivsMed. We note that alternative wristed instruments also exist. While the kinematic modeling described below depends on the specific tool geometry, the enforcement of the RCM constraint is formulated to remain instrument-agnostic.

Compared with fully actuated robotic instruments, incorporating manually-wristed laparoscopic tools into robotic teleoperation from a humanoid robot involves a substantially more passive kinematic chain, since these tools are intended for direct human operation and do not provide actively actuated joints. They furthermore rely on human visuo-haptic hand-eye coordination of a remote center of motion, which must now be mimicked by a humanoid robot. Consequently, the overall mechanism must be reconstructed from the relative poses between components. We express all instrument frames in the robot base coordinates (Fig.\ref{fig:IK}), and treat the robot end-effector, grasper, and ``handle 1'' frames as a single rigid body without internal motion.

We parameterize the passive tool geometry using $\theta_1,\theta_2$ for the relative handle angles (w.r.t.\ neutral), and $\theta_3,\theta_4$ for the corresponding tool-tip articulation angles. Given the humanoid joint configuration $\boldsymbol{q}$, the end-effector (wrist) pose is obtained by forward kinematics,
\begin{equation}
    \textbf{P}_{ee} = f_{FK}(\boldsymbol{q}) \in \mathbb{R}^{4 \times 4},
\end{equation}
and the handle~1 pose follows from a fixed transform ${}^{ee}\textbf{T}_{h1}$ between the wrist and handle~1:
\begin{equation}
    \textbf{P}_{h1} = \textbf{P}_{ee}{}^{ee}\textbf{T}_{h1},
\end{equation}
with ${}^{ee}\textbf{T}_{h1} \in \mathbb{R}^{4 \times 4}$ being constant.

The RCM position $\textbf{x}_{rcm} \in \mathbb{R}^3$ is calibrated using ArUco markers mounted on the laparoscopy trainer board. Let $l_{0}$ denote the distance from handle~2 to the RCM, and $l_{12}$ the separation between handles~1 and~2, which are measured directly from the tool. Here $\theta_1$ is computed as
\begin{equation}
    \theta_1 = \arccos\left(\frac{l_{12}^2 + l_0^2 - \|\textbf{x}_{rcm} - \textbf{x}_{h1}\|^2}{2l_{12}l_0}\right).
\end{equation}
The passive linkage satisfies
\begin{equation}
    \|\textbf{x}_{h1} - \textbf{x}_{h2}\| = l_{12}, \quad
    \|\textbf{x}_{rcm} - \textbf{x}_{h2}\| = l_{0},
\end{equation}
together with the perpendicularity constraint
\begin{equation}
    (\textbf{R}_{h1}[0,0,1]^T)\cdot(\textbf{x}_{h1}-\textbf{x}_{h2})^T=0.
\end{equation}
Solving these relations provides us with the handle~2 position
\begin{equation}
    \textbf{x}_{h2} = \textbf{x}_{h1} + l_{12} \cdot \frac{\textbf{n}_1 \times (\textbf{n}_1 \times \textbf{n}_2)}{\|\textbf{n}_1 \times (\textbf{n}_1 \times \textbf{n}_2)\|},
\end{equation}
where
\begin{equation}
    \textbf{n}_1 = \textbf{R}_{h1}[0,0,1]^T, \quad \textbf{n}_2 = \textbf{x}_{rcm} - \textbf{x}_{h1}.
\end{equation}
From handle~2 to the tool tip, the remaining segment is modeled as a rigid link of known length $l_1$:
\begin{equation}
    \textbf{x}_{tt} = \textbf{x}_{h2} + \frac{l_1}{l_0}(\textbf{x}_{rcm}-\textbf{x}_{h2}).
\end{equation}
With $\textbf{x}_{h2}$ determined, $\theta_2$ is computed by
\begin{equation}
    \theta_2 = \arccos\left(\frac{\textbf{v}_1 \cdot \textbf{v}_2}{\|\textbf{v}_1\| \cdot \|\textbf{v}_2\|}\right),
\end{equation}
where
\begin{equation}
    \textbf{v}_1 = \textbf{x}_{h1}-\textbf{x}_{h2}, \quad 
    \textbf{v}_2 = \textbf{R}_{h1}[1,0,0]^T \; .
\end{equation}

The tool-tip orientation is then reconstructed as
\begin{equation}
    \textbf{R}_{tt} = \textbf{R}_{h1}\,\textbf{R}(-\theta'_2)\,\textbf{R}(-\theta'_1)\,\textbf{R}(\theta'_3)\,\textbf{R}(\theta'_4),
\end{equation}
with the coupled tip angles given by
\begin{equation}
    \theta_3 = k \theta_1, \quad
    \theta_4 = k \theta_2,
\end{equation}
where the scaling factor $k=2$ is identified as the gearing ratio of the tool's internal mechanism. The signed angles $\theta'_1$ and $\theta'_2$ are defined as
\begin{equation}
    \theta'_1 = \frac{(z_{h1}-z_{rcm})}{\|z_{h1}-z_{rcm}\|}\cdot \theta_1, \quad
    \theta'_2 = \frac{(\textbf{v}_1 \times \textbf{v}_2)\cdot \textbf{n}_2}{\|(\textbf{v}_1 \times \textbf{v}_2)\cdot \textbf{n}_2\|}\cdot \theta_2
\end{equation}
Finally, the tool tip pose is written as
\begin{equation}
    \mathbf{P}_{tt} =
    \begin{bmatrix}
    \mathbf{R}_{tt} & \mathbf{t}_{tt}\\
    \mathbf{0}^\top & 1
    \end{bmatrix}.
\end{equation}

This model captures the extended passive chain of manually-wristed laparoscopic instruments, enabling recovery of both tool-tip position and orientation. Given a target tool-tip pose $\mathbf{P}_{tt}' \in SE(3)$ and the RCM location $\mathbf{x}_{\text{rcm}}$, we solve for the handle~1 pose $\mathbf{T}_{h1} \in SE(3)$ by minimizing a weighted residual while enforcing soft joint angle limits. The residual, combining pose mismatch and passive angle penalties, is defined as
\begin{equation}
\mathbf{r}(\mathbf{P}_{h1}) =
\begin{bmatrix}
w_t \big( \mathbf{x}_{tt} - \mathbf{x}_{tt}' \big) \\[4pt]
\mathrm{log}\!\left(\mathbf{R}_{tt}^{\top} \mathbf{R}_{tt}'\right) \\[4pt]
w_a \cdot \max(0,|\theta_1|-\theta_{\max}) \\[2pt]
w_a \cdot \max(0,|\theta_2|-\theta_{\max}\big)
\end{bmatrix},
\label{eq:ik_residual}
\end{equation}
where $\mathbf{x}_{tt}$ and $\mathbf{R}_{tt}$ are computed from the current $\mathbf{P}_{h1}$. Here, $w_t$ and $w_a$ weight the translation and angle-limit terms, and $\theta_{\max}$ is set to 45 degrees to reflect the mechanical constraint. We obtain the inverse-mapped handle~1 pose using the Trust-Region Reflective (TRF) algorithm with a linear loss:
\begin{equation}
\mathbf{P}_{h1}^\star = \arg\min_{\mathbf{P}_{h1}} \;\; \mathcal{L}\!\big(\mathbf{r}(\mathbf{P}_{h1})\big) \; .
\end{equation}
The corresponding target robot end-effector pose $\mathbf{P}_{ee}^\star$ is then computed by applying the physical offset $\textbf{t}_{offset}$ from the handle~1 frame to the robot wrist frame:
\begin{equation}
    \mathbf{P}_{ee}^\star = \mathbf{P}_{h1}^\star \begin{bmatrix}
    \mathbf{I} & \mathbf{t}_{offset}\\
    \mathbf{0}^\top & 1
    \end{bmatrix} \; .
\end{equation}



\backmatter

\section{Supplementary information}
Supplementary video 1 includes a demonstration of the humanoid teleoperation system, a dry lab
user study session, and an overview of the two live surgeries conducted. Supplementary video 2 covers detailed procedures for the live porcine laparoscopic cholecystectomies.


\section{Acknowledgments}

We thank our labmates in the Advanced Control and Robotics Lab (ARC Lab) for their assistance with hardware transfer. We also thank the staff of the Center for the Future of Surgery (CFS) at the University of California, San Diego, for coordinating the experimental space and facilities.



\section*{Declarations}




\subsection{Conflict of interest/Competing interests} 
S.L. is a Principal Investigator of the Alume Trial at UC San Diego, R.B. is a consultant to Stryker, Johnson \& Johnson MedTech, and DistalMotion. M.Y. is co-Founder and Board Member of Channel Robotics and Owner-Operator of Yip Consulting Services.

\subsection{Ethics approval and consent to participate}
All procedures were conducted with institutional animal care approval (UCSD IACUC \#S08027). A licensed veterinarian administered and monitored the animals’ general anesthesia, while simulation center staff assisted with sterile preparation and draping of the animal before incision.



\subsection{Code/data availability}
 The code supporting the findings of this study is publicly available on Zenodo (DOI: 10.5281/zenodo.18023650).
\subsection{Author contribution}
Methodology: Z.L., P.Z., M.Y.; Study design: Z.L., N.T., M.Y.; Experiments: Z.L., N.T., P.Z., C.J, S.A., G.J., S.L., R.B., M.Y.; Data analysis: Z.L., N.T.; Paper writing: Z.L., N.T., P.Z.,  F.R., M.Y.; Conceptualization: S.L., R.B., M.Y.

\bibliography{sn-bibliography}

\end{document}